\documentclass[10pt,twocolumn]{article}

\usepackage{arxiv}

\usepackage[utf8]{inputenc} 
\usepackage[T1]{fontenc}    
\usepackage{hyperref}       
\usepackage{url}            
\usepackage{booktabs}       
\usepackage{nicefrac}       
\usepackage{microtype}      

\usepackage{amsmath,amssymb,amsfonts,bm}
\usepackage{algorithmic}
\usepackage{eurosym} 
\usepackage[export]{adjustbox} 
\usepackage{graphicx}
\usepackage{float}  
\usepackage{filecontents} 
\usepackage{caption} 
\usepackage{subcaption} 
\usepackage{dsfont}  
\usepackage{textcomp}
\usepackage{epsfig} 
\usepackage{xcolor}
\usepackage{tikz}
\usepackage[numbers,sort]{natbib}

\newcommand*\samethanks[1][\value{footnote}]{\footnotemark[#1]}

\newcommand\copyrighttext{%
  \footnotesize {\large IEEE Copyright Notice}\\
\textcopyright 2020 IEEE.  Personal use of this material is permitted.  Permission from IEEE must be obtained for all other uses, in any current or future media, including reprinting/republishing this material for advertising or promotional purposes, creating new collective works, for resale or redistribution to servers or lists, or reuse of any copyrighted component of this work in other works.}
\newcommand\copyrightnotice{%
\begin{tikzpicture}[remember picture,overlay]
\node[anchor=south,yshift=10pt] at (current page.south) {\fbox{\parbox{\dimexpr\textwidth-\fboxsep-\fboxrule\relax}{\copyrighttext}}};
\end{tikzpicture}%
}

\title{ActGAN: Flexible and Efficient One-shot Face Reenactment}

\author{
  Ivan Kosarevych\thanks{These two authors contributed equally} \textsuperscript{ ,1, 2},
  Marian Petruk\samethanks\hspace{0.05cm} \textsuperscript{,1, 2}, \\ Markian Kostiv\textsuperscript{1, 2}, Orest Kupyn\textsuperscript{2, 3},
Mykola Maksymenko\textsuperscript{1},
Volodymyr Budzan\textsuperscript{1, 3}\\
\textsuperscript{1}\textit{SoftServe, Inc.:}
{\texttt{\{ikosar, mpetru, mkost, mmaks\}@softserveinc.com}}\\
\textsuperscript{2}\textit{Ukrainian Catholic University:}
{\texttt{\{kosarevych, petruk, m.kostiv, kupyn\}@ucu.edu.ua}}\\
\textsuperscript{3}{\texttt{\{orestkupyn, budzan.v\}@gmail.com}}
}

\begin{document}



\maketitle

\begin{abstract}
This paper introduces ActGAN -- a novel end-to-end generative adversarial network (GAN) for one-shot face reenactment.
Given two images, the goal is to transfer the facial expression of the source actor onto a target person in a photo-realistic fashion.
While existing methods require target identity to be predefined, we address this problem by introducing a \textit{``many-to-many''} approach, which allows arbitrary persons both for source and target without additional retraining.
To this end, we employ the Feature Pyramid Network (FPN) as a core generator building block -- the first application of FPN in face reenactment, producing finer results.
We also introduce a solution to preserve a person's identity between synthesized and target person by adopting the state-of-the-art approach in deep face recognition domain.
The architecture readily supports reenactment in different scenarios: \textit{``many-to-many''}, \textit{``one-to-one''}, \textit{``one-to-another''} in terms of expression accuracy, identity preservation, and overall image quality.
We demonstrate that ActGAN achieves competitive performance against recent works concerning visual quality.
\end{abstract}

\begin{figure}[!b]
\vspace{7.5cm}
        \includegraphics[width=1.0\linewidth,height=3.2cm, keepaspectratio]{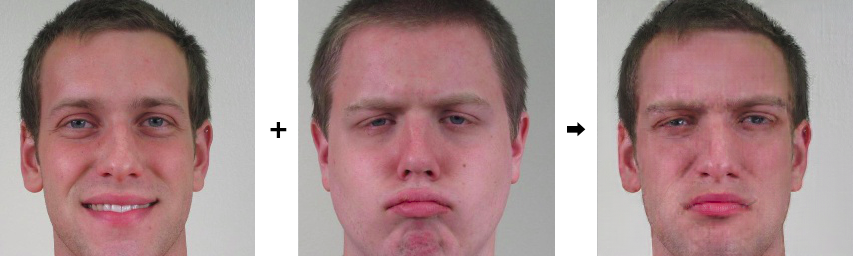}
        \includegraphics[width=1.0\linewidth,height=3.2cm,keepaspectratio]{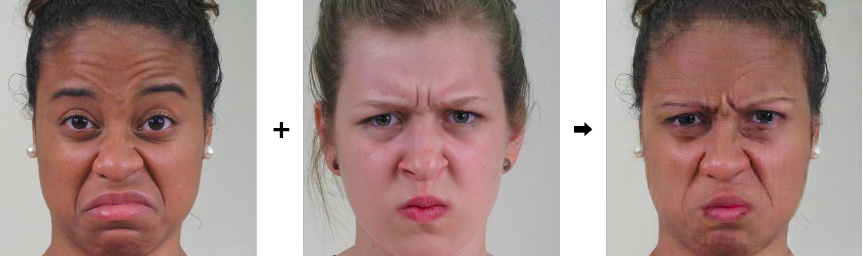}
        \includegraphics[width=1.0\linewidth,height=3.2cm,keepaspectratio]{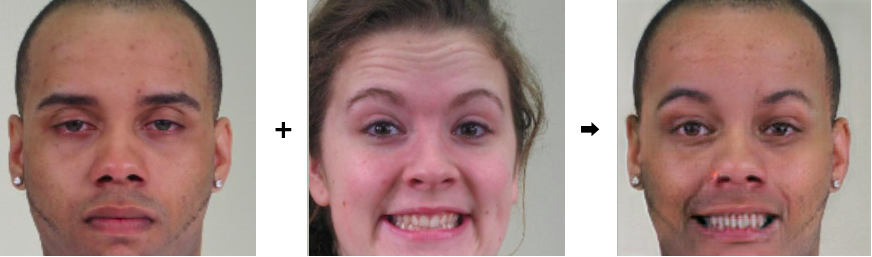}
        \includegraphics[width=1.0\linewidth,height=3.9cm,keepaspectratio]{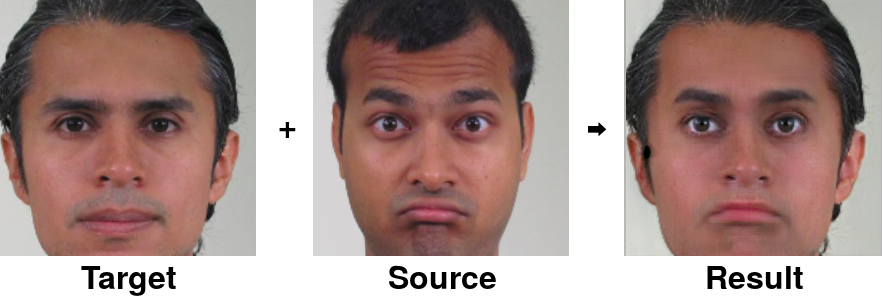}
  \caption{
    Examples of face reenactment using the proposed model. 
    On the left side of each image triplet, one may see an input image of the target, whose identity and background we aim to preserve.
    On the right side, a source's input image, whose facial expression we use to modify the target.
    In the middle, one may see the output of the generator - reenacted face of the target with expression from the source.
}
\label{fig:multi2600}
\end{figure}


\section{Introduction}
\copyrightnotice
Face reenactment aims at transferring of facial expression from source to target~\cite{2018arXiv180711079W, Thies:2015:RET:2816795.2818056}.
This field of research has a myriad of real-world applications, such as cinematography, mass media, AR/VR, computer games, telepresence, to name a few.

The most challenging part of reenactment consists in the preservation of target identity, addressed in several recent papers~\cite{2019arXiv190508233Z, ha2019marionette, 2019arXiv190511805Z, 2018arXiv180709251P}.
The problem grows in complexity if we aim to preserve the identity of \textit{unseen} targets.
Rule-based approaches to identity preservation may fail due to the need for a high number of handcrafted parameters.\clearpage

The appearance of Generative Adversarial Network (GAN)~\cite{2014arXiv1406.2661G} based approaches in image-to-image transfer~\cite{2016arXiv161107004I, 2017arXiv171111585W, 2017arXiv170310593Z, 2017arXiv170300848L} introduces a captivating alternative for face reenactment.

Therefore recent studies \cite{Kim:2018:DVP:3197517.3201283, 2019arXiv190508233Z, 2018arXiv180711079W, 2017arXiv171006090X, 2018arXiv180709251P, tripathy2019icface} adopt adversarial training.

Following them, we also exploit GAN in our system.
However, existing methods such as \cite{2018arXiv180711079W, Kim:2018:DVP:3197517.3201283} are limited in scalability when a target person is not predefined.
Such methods have to train large networks, where both generator and 
discriminator have a vast number of parameters for each target person.
Meanwhile, solutions that overcome this limitation, such as Zakharov \emph{et al.}~\cite{2019arXiv190508233Z}, require additional retraining.

We tackle the above problems with the following novelties:

\begin{itemize}
    \item the ability to generate reenacted faces in \textit{``many-to-many''} scenario in a \textit{one-shot} manner; 
    \item the solution based on the state-of-the-art model in the face-recognition domain -- ArcFace~\cite{2018arXiv180107698D} to preserve a person's identity between synthesized and the target person;
    \item inspired by DeblurGAN-v2~\cite{2019arXiv190803826K} we pioneer usage of Feature Pyramid Network~\cite{2016arXiv161203144L} to face reenactment, this allows us to create an efficient, compact in size model;
    \item the flexible structure of the proposed end-to-end network framework that is agnostic to the choice of the generator for the balance between generation quality and efficiency.
\end{itemize}

\section{Related Work}

Face-reenactment has the following main objectives: \mbox{1) facial} expression transfer; \mbox{2) identity} preservation; \mbox{3) background} and illumination retention.
Apart from these, others may be considered, such as head pose or eye gaze transfer~\cite{Kim:2018:DVP:3197517.3201283, 7780631, Thies:2015:RET:2816795.2818056}.

One of the ways to represent a facial expression is to model it with landmarks, spread along the face in 2D space \cite{2018arXiv180711079W, 2019arXiv190508233Z}. 
Another way is a 3D face mask, which models face more accurately, but is more computationally expensive \cite{Thies:2015:RET:2816795.2818056, Kim:2018:DVP:3197517.3201283}.

The secondary objective, \emph{i.e.} Face identities of the target per\-son and a generated one should be the same.
To this end, Kim \emph{et al.}~\cite{Kim:2018:DVP:3197517.3201283} use dense face reconstruction that fits a para\-met\-ric model of face and illumination, while Zakharov \emph{et al.}~\cite{2019arXiv190508233Z} apply additional fine-tuning on the specific target person.

In Wang \emph{et al.}~\cite{2019arXiv190601314W} synthetic background may be blurry, while we aim at conserving the background (third objective) by applying perceptual loss~\cite{2016arXiv160308155J} between synthesized and target images. 

\section{Methods}

The important aspects of our approach include generator architecture, discriminator's behaviour, face normalization and identity mismatch calculation, which are described in more details in the corresponding sections below. 

In recent studies, Feature Pyramid Network framework~\cite{2016arXiv161203144L} shows impressive results, \emph{e.g.} in object detection and segmentation~\cite{2019arXiv190102446K}.
Rich feature extraction and reconstruction abilities of FPN allow us to blend images of source and target carefully.

We exploit the benefits of conditional discriminator by conditioning on facial landmarks.
Instead of calculating mismatch between generated and target expressions directly, we make discriminator distinguish between a fake and a real expression (details in section~\ref{net_architecture}) similarly to~\cite{DBLP:journals/corr/abs-1812-08861}.
It gives extra freedom to a generator.
In such conditions, the generator is not supposed to produce expression exactly, but rather is taking into account the personal characteristics of a target.

In order to provide stable working of the proposed system, face normalization is needed (\ref{face_normalization}), \emph{i.e.} source and target faces should be similarly aligned on the image. 

The normalization process requires source and target to have similar head poses, therefore we consider only frontal faces.

\subsection{Network Architecture}
\begin{figure*}[!ht]
	\centering
    \makebox[0pt]{\includegraphics[width=0.8\paperwidth,height=4.5cm,keepaspectratio]{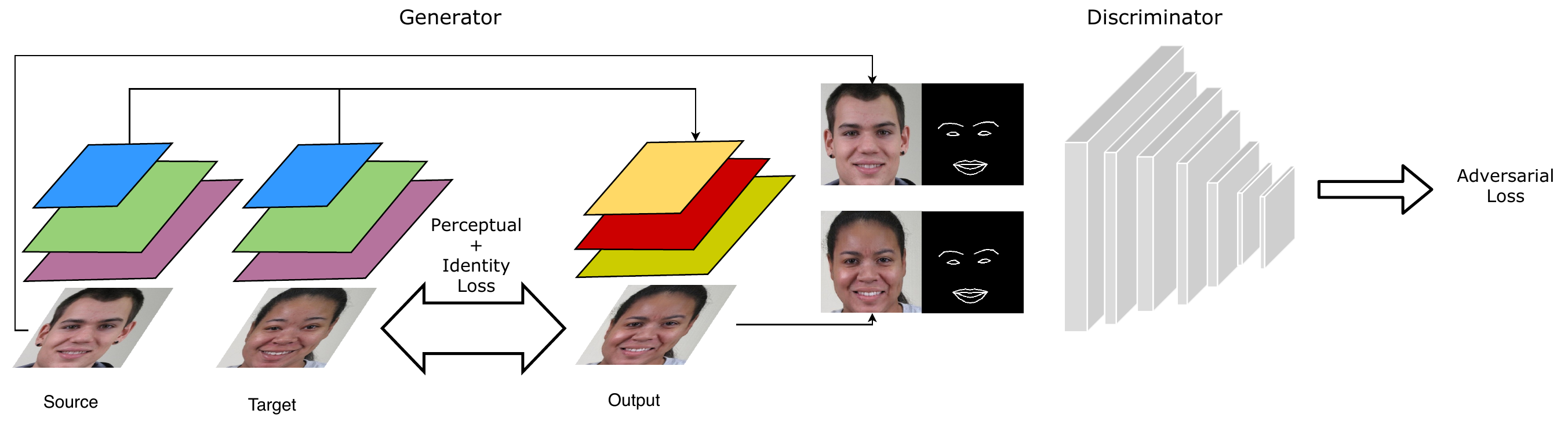}}
    \caption{
        High-level architecture diagram.
        Images of source and target are propagated through Generator (Fig.~\ref{fig:generator_diagram}).
        Computed face landmarks of source stacked with both source and generated images fed into the Discriminator. 
        It computes adversarial loss, which forces the generator to produce more accurate expression.
        Also, between target and generated image, we compute perceptual~\cite{2016arXiv160308155J} (briefly described in~\ref{net_train}) and identity loss (described in~\ref{face_identity_loss}) for better content and identity preservation.
    }
    \label{fig:network_architecture_diagram}
\end{figure*}

\begin{figure*}[!ht]
	\centering
    \makebox[0pt]{\includegraphics[width=0.9\paperwidth,height=5cm,keepaspectratio]{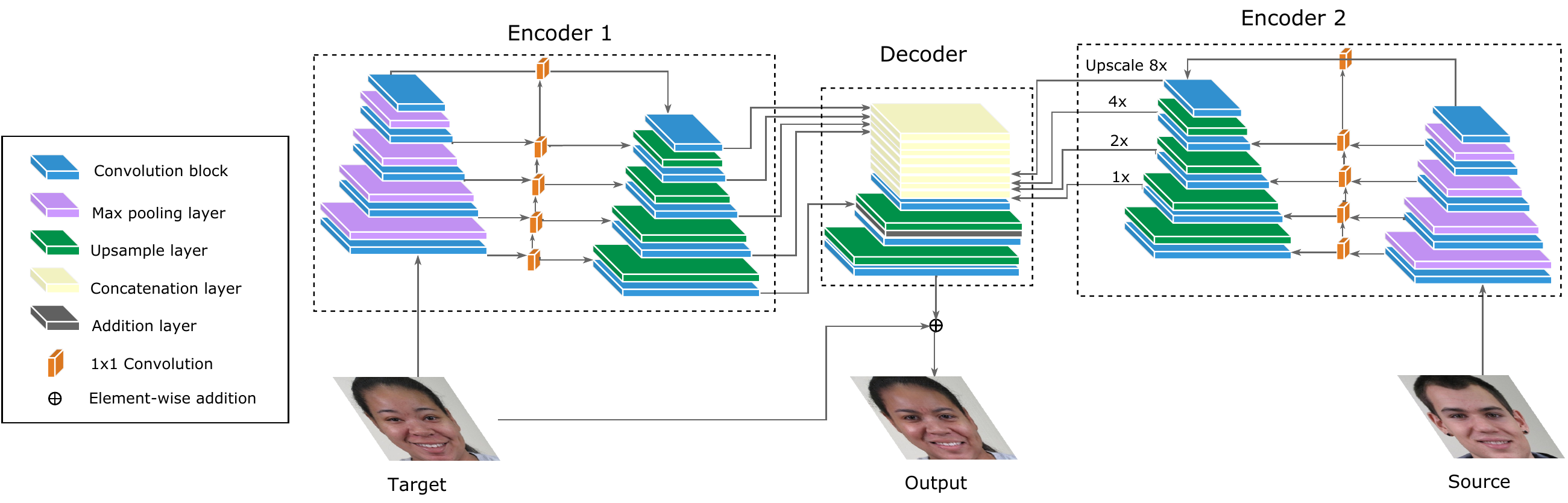}}
    \caption{
        High-level FPN-based Generator architecture diagram.
        Both source and target images fed into separate encoders, they are organized in the Feature Pyramid Network.
        In FPN on the bottom-up path, five feature maps (five semantic levels) are extracted and propagated through 1\texttimes1 convolution.
        The topmost convolution is the beginning of the top-down path. 
        Every feature map from the bottom-up path is then upsampled and added to the corresponding map from the top-down pathway.
        Finally, we receive five feature maps both from source and target Encoders.
        In the Decoder the top four of these maps are upscaled, concatenated and propagated through a series of layers (convolutional and upsampling).
        The bottom-most feature map from the target's Encoder added after first upsampling.
        In the end, the target image is added to the result of the Decoder to get generated the face.
    }
    \label{fig:generator_diagram}
\end{figure*}

The proposed pipeline (Fig.~\ref{fig:network_architecture_diagram}) uses a standard adversarial setup with a single generator (Fig.~\ref{fig:generator_diagram}) and a discriminator.

\label{net_architecture}

\subsubsection{Generator} \label{generator}

Our generator is designed in an end-to-end fashion.
It infers how to reconstruct the desired image $\hat{x}$ straight from the source $x$ and target $x\prime$ pictures.

It is organized in an encoder-decoder way and based upon Feature Pyramid Network~\cite{2016arXiv161203144L}.

FPN combines features from different semantic levels, \emph{i.e.} more or less detailed features, extracted on the bottom-up path with usual convolutional layers, to enrich spatial reconstruction on a top-down pathway via lateral connections.

For feature extraction by default we took InceptionResNetV2\- with weights pre-trained on ImageNet. 
The combination we use allows exploiting all the benefits of the residual approach while retaining the computational efficiency of Inception network~\cite{2016arXiv160207261S}.
This approach may adjust to other ConvNets.

The generator consists of two encoders with the same FPN-based architecture.
One of them learns the distribution of $x$, and the other one the distribution of $x\prime$.
In this way, we do not mix up features extracted from $x$ and $x\prime$ before decoding.

Extracted information from different scales is stored in two sets of maps for $x$ and $x\prime$ respectively.
Within the decoding part, we employ a sequence of convolutions and upsamplings on the maps from both sets concatenated altogether.
It allows the model to learn information on different semantic levels from both of encoded images simultaneously.

\subsubsection{Discriminator}
The discriminator is a five-layer fully-convolutional network.
Hidden layers use InstanceNorm~\cite{ulyanov2016instance}, which is preferred over BatchNorm~\cite{ioffe2015batch} for GANs.
Neurons in layers get activated with LeakyReLU, that prevents the appearance of ``dying neurons''~\cite{lu2019dying}.
Discriminator outputs a single number, following the idea in Relativistic GANs~\cite{2018arXiv180700734J}.

\subsection{Face Landmarks Estimation}
\vspace{-0.1cm}

We project an image onto landmark latent space, which holds adequate geometry information, being at the same time indifferent to identity.
Face landmarks are represented as a set of $N$ ($x$,~$y$) coordinates on $H \times W$ image.
They are calculated using Dlib~\cite{dlib09} twice in our pipeline.
First, for face normalization during the dataset preparation stage (described in subsection~\ref{face_normalization}).
Second, we exploit landmarks for network training (subsection~\ref{net_train}), where they come as a part of discriminator input.
We interpolate landmarks to get dense boundary lines before feeding them into discriminator following the idea in~\cite{2018arXiv180711079W}.

\subsection{Face normalization}
\label{face_normalization}

There is a significant structural gap between the face shapes of source and target, which may lead to severe artefacts~\cite{2018arXiv180711079W}.
We do normalization (alignment) to make landmarks of different images appear similar to a predefined configuration.

This operation is performed on the original image, meaning we do not crop face region beforehand.
The process works as follows; first, we select five facial points, namely eyes, nose, and two mouth corners. 
Then, these points are adopted to perform similarity transformation.
Finally, we obtain a cropped face, which is then resized to $N \times N$ region, where $N=256$ px. ($N=512$ px. also provides plausible results, albeit with more artefacts).

\subsection{Face Identity Loss} \label{face_identity_loss}

To represent a person's identity, we encode it in a vector of features.
In order to extract such identity embedding from the image, we adopt Additive Angular Margin Loss (ArcFace) proposed by Deng \emph{et al.}~\cite{2018arXiv180107698D}, which is a state-of-the-art model in the face-recognition domain.
In our experiments, we use the pre-trained ArcFace model with Squeeze-and-Excitation ResNet-50~\cite{2017arXiv170901507H} backbone.
The authors of the aforementioned paper Jie Hu \emph{et al.} show that by adding SE-blocks to ResNet-50, one can expect almost the same accuracy as ResNet-101 delivers.
This way, we need less computational resources to obtain higher accuracy.  

Finally, to find how much identity of $\hat{x}$ varies from $x\prime$ we evaluate the distance between corresponding calculated embeddings $E_{\hat{x}}$ and $E_{x\prime}$.
\begin{equation}
    L_{\text{identity}} = \sum{(E_{\hat{x}} - E_{x\prime})^2}
\end{equation}

\subsection{Network Training}
\label{net_train}
We train our model on target-source pairs of images selected at random.

We feed a pair per propagation because we adopt Instance normalization~\cite{ulyanov2016instance}.
Such methodology, by our observations, provides more stable training.

Studies show that instance normalization performs well on different tasks such as style transfer or dehazing~\cite{2018arXiv180503305X} as an alternative to batch normalization.
Therefore it is now commonly used to replace the BatchNorm~\cite{ioffe2015batch} in GANs.

In order to achieve realistic facial reenactment, we make generator pursue three objectives.

The proposed identity loss $L_{\text{identity}}$ described in~\ref{face_identity_loss} to prevent generator from modifying face characteristics of a target person.
Illumination and background changes are constrained with the perceptual loss $L_{\text{content}}$, which is $L2$ norm in activations of the third convolutional layer of pre-trained VGG19 model between $x\prime$ and $\hat{x}$~\cite{2016arXiv160308155J}.

Stable and effective training of discriminator is achieved by applying RaGAN-LS loss function from the family of
Relativistic average GANs (RaGANs), which generate higher quality data than non-reltivistic ones~\cite{2018arXiv180700734J}.  
Therefore adversarial loss $L_{\text{adv}}$ is as following: \begin{multline}
    L_{\text{adv}} = \mathrm{L}_G^{\text{RaLSGAN}} =\\ \mathds{E}_{x_\gamma}[(D(x_\gamma) - \mathds{E}_{(x, x^\prime)} D(G_\gamma(x, x^\prime))+1)^2] \\ + \mathds{E}_{(x, x^\prime)}[(D(G_\gamma(x, x^\prime)) - \mathds{E}_{x_\gamma} D(x_\gamma)-1)^2]
 \end{multline} where $x_\gamma$ - source image $x$ concatenated with its face landmarks $\gamma$; $G_\gamma(x, x^\prime)$ - generated image concatenated with source's face landmarks $\gamma$.
Full objective combines three losses named above with appropriate scales:
\begin{equation} \label{main-objective}
    L_{\text{total}} = \lambda_{\text{content}} \times L_{\text{content}} + \lambda_{\text{adv}} \times L_{\text{adv}} + \lambda_{\text{identity}} \times L_{\text{identity}}
\end{equation}

\section{Experiments}
\label{experiments}
We compare our model with the state-of-the-art approaches, as well as run ablation study of different model architectures and training methods.
We evaluate three components, which are significant for face reenactment: 1) image realism, 2) expression accuracy, 3) identity preservation. 

\begin{table}
    \centering
	\begin{tabular}{lcccc}
		\toprule
		
		 & FID $\downarrow$ & NMSE $\downarrow$ & CSIM $\uparrow$ \\
		\midrule
		
		& \multicolumn{3}{c}{\textit{``many-to-many''}} \\ \cmidrule(lr){2-4}
        (a) & $45.01$ & \bm{$4.19 \%$} & $0.45$ \\ 
		\addlinespace
	    (b) & \bm{$29.97$} & $5.04 \%$ & \bm{$0.85$} \\
		\bottomrule
	\end{tabular}
	\caption{Ablation study for siamese vs separate encoders in generator architecture.
	(a)  -  siamese encoders;
	(b)  -  separate encoders.}
	\label{tab-encoders_ablation}
\end{table}

Results in these three components may vary depending on different reenactment scenarios.
We explored three possible scenarios.
For every scenario, we created the corresponding dataset.
\begin{enumerate}
  \item \textit{``Many-to-many''} - source and target identities and expressions are different, randomised (\emph{i.e.} many identities) in the dataset;
  \item \textit{``One-to-one''} - source and target identities are the same, facial expressions are different;
  \item \textit{``One-to-another''} - source identity is different from a target one, but their identities are constant in the dataset, different expressions.
\end{enumerate}

The first case is the one in which we are most interested.
It is the most challenging so far, mainly in identity and background preservation.
Within the second scenario, we study whether self-reenactment, \emph{i.e.} into the same person, can improve the overall quality of the image and modify a person's identity less than in a \textit{``many-to-many''} case.
Finally, the third case is an intermediate one between the first two.

\subsection{Implementation Details}
Experimentaly the following coefficients produced the most satisfactory results: $\lambda_{\text{content}}=0.01$, $\lambda_{\text{adv}}=0.001$, $\lambda_{\text{identity}}=0.001$ (for Formula \ref{main-objective}).

For this research, we are using Pytorch~\cite{2019arXiv191201703P}.
The proposed model and all the models for ablation study are trained from scratch on an NVIDIA GTX 1080 GPU using the Adam solver~\cite{2014arXiv1412.6980K} with a batch size of 1.
The learning rate is initially set to $0.0001$ with linear decay down to \(1\mathrm{e}{-7}\) starting from 40 epoch.

\subsection{Datasets}
For training, we selected reduced EmotioNet dataset~\cite{7780969}, which has 26 different emotions represented by more than 200 persons.
Dataset was organized in \textit{``many-to-many''} manner, meaning for every target person, the source one is any other person with distinct identity and expression.

For evaluation and comparisons, we randomly selected samples distinct from training ones.

To preprocess the data, we do face normalization, explained in~\ref{face_normalization}.

\subsection{Evaluation Metrics}
We evaluate our method on standard metrics connected to image quality, realism and expression transfer accuracy, namely: FID, NMSE, CSIM.

We apply Fr\'echet inception distance (FID)~\cite{2017arXiv170608500H} to measure the variation and realism of generated images.

For semantic evaluation, \emph{i.e.} the correspondence between the source's landmarks and the landmarks on the synthesized image, we employ NMSE (normalized by inter-ocular (centroid of an eye) distance mean squared error (times 100\%)) commonly used in many~\cite{DBLP:journals/corr/abs-1812-03887, Cao:2014:FAE:2597520.2597529, 6619290, DBLP:journals/corr/ZhangLLT14, 2016arXiv160301249R, 2017arXiv171100253C} papers to compare semantic information \emph{i.e.} face expression through interpolated landmarks.

\begin{equation}
    \textsc{NMSE} = \frac{\sum\limits_{i=1}^{L} \sqrt{(x^\prime_i - \hat{x}_i)^2 + (y_i^\prime - \hat{y}_i)^2}}{L \cdot \sqrt{(x^\prime_l - x^\prime_r)^2 + (y^\prime_l - y^\prime_r)^2}} \cdot 100
\end{equation}
where $L$ - number of landmarks, $x^\prime_l$ - x-coordinate of left pupil of the source (ground truth), $y^\prime_l$ - y-coordinate of left pupil of the source, similarly $x^\prime_r$ and $y^\prime_r$ - coordinates of the right pupil.

To compare identity preservation of the generated image, we use CSIM metric - cosine similarity between embedding vectors of ArcFace model~\cite{2018arXiv180107698D}.

\begin{table}
    \centering
	\begin{tabular}{lccc}
		\toprule
		
		 & FID $\downarrow$ & NMSE $\downarrow$ & CSIM $\uparrow$ \\
		\midrule
		
		& \multicolumn{3}{c}{\textit{``many-to-many''}} \\ \cmidrule(lr){2-4}
        (a) & $63.43$ & $6.83 \%$ & $0.24$ \\ 
		\addlinespace
	    (b) & \bm{$29.97$} & \bm{$5.04 \%$} & \bm{$0.85$} \\
		\bottomrule
	\end{tabular}
	\caption{Quantitative comparison of pix2pixHD vs our proposed model in \textit{``many-to-many''} reenactment scenario.
	(a) - pix2pixHD;
	(b) - Our (Full objective FPNInceptionResNet with separate encoders).}
	\label{tab-pix2pix_ablation}
\end{table}

\begin{table}
    \centering
	\begin{tabular}{cc|cc}
		\toprule
		
		(a) & (b) & (c) & (d) \\
		\midrule
		\multicolumn{4}{c}{\textit{``many-to-many''}} \\ \cmidrule(lr){1-4}
		
        \includegraphics[width=1.6cm, height=1.6cm]{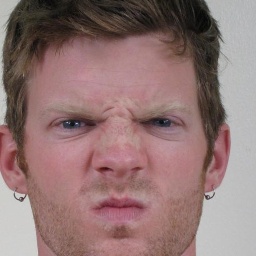} &
	    \includegraphics[width=1.6cm, height=1.6cm]{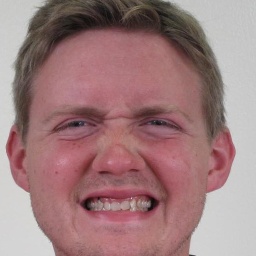} &
	    \includegraphics[width=1.6cm, height=1.6cm]{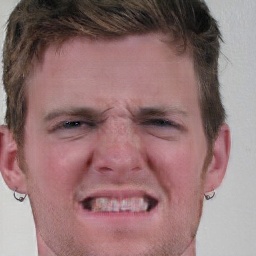} 
	    &
	    \includegraphics[width=1.6cm, height=1.6cm]{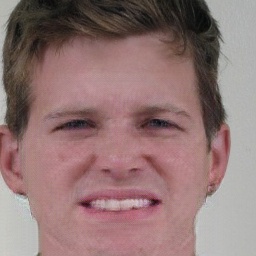}\\
	    
	    \includegraphics[width=1.6cm, height=1.6cm]{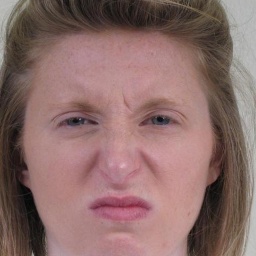} &
	    \includegraphics[width=1.6cm, height=1.6cm]{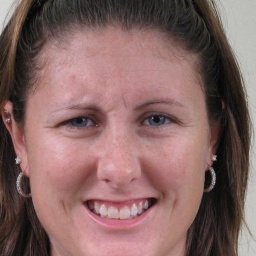} &
	    \includegraphics[width=1.6cm, height=1.6cm]{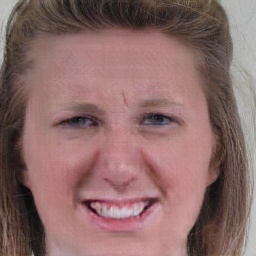} & \includegraphics[width=1.6cm, height=1.6cm]{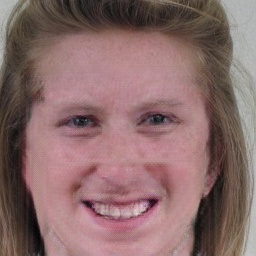} \\
	    
	    \includegraphics[width=1.6cm, height=1.6cm]{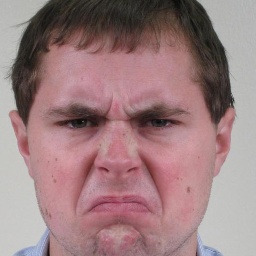} &
	    \includegraphics[width=1.6cm, height=1.6cm]{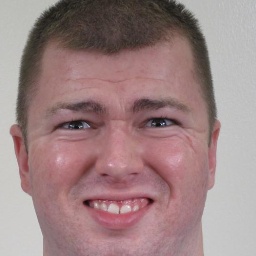} &
	    \includegraphics[width=1.6cm, height=1.6cm]{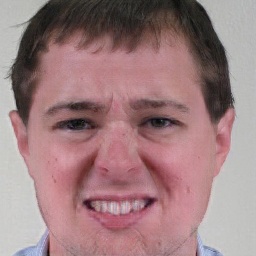} & 
	    \includegraphics[width=1.6cm, height=1.6cm]{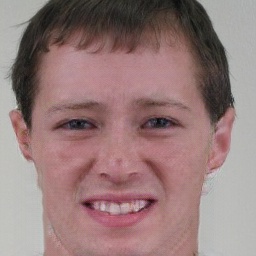} \\
		\bottomrule
	\end{tabular}
	\caption{Qualitative comparison of our proposed model vs pix2pixHD.
	(a) - target image;
	(b) - source image;
	(c) - Our model;
	(d) - pix2pixHD.}
	\label{tab-pix2pix-ablation-img}
\end{table}

\begin{table}
    \centering
	\begin{tabular}{cc|cc}
		\toprule

		(a) & (b) & (c) & (d) \\
		\midrule
		\multicolumn{4}{c}{\textit{``many-to-many''}} \\ \cmidrule(lr){1-4}
		
		\addlinespace
	    
	    \includegraphics[width=1.6cm, height=1.6cm]{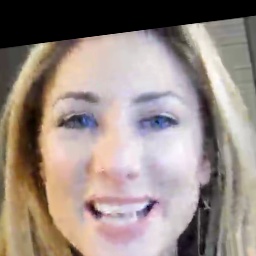} &
	    \includegraphics[width=1.6cm, height=1.6cm]{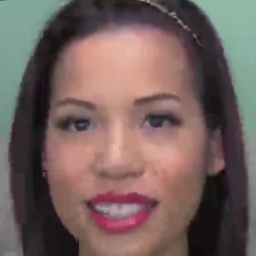} &
	    \includegraphics[width=1.6cm, height=1.6cm]{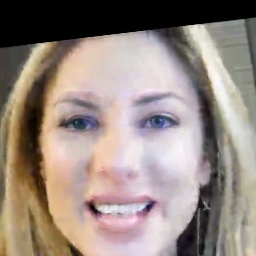} &
	    \includegraphics[width=1.6cm, height=1.6cm]{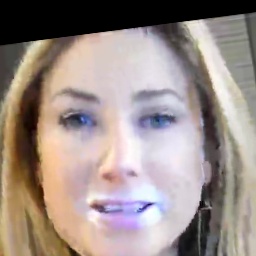} \\
	    
	    
	    \includegraphics[width=1.6cm, height=1.6cm]{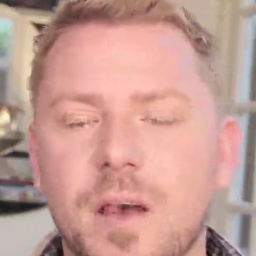} &
	    \includegraphics[width=1.6cm, height=1.6cm]{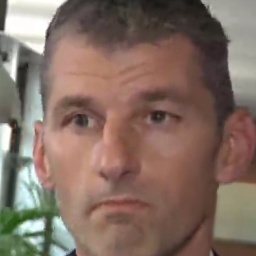} &
	    \includegraphics[width=1.6cm, height=1.6cm]{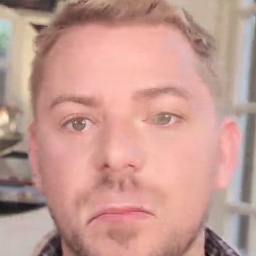} &
	    \includegraphics[width=1.6cm, height=1.6cm]{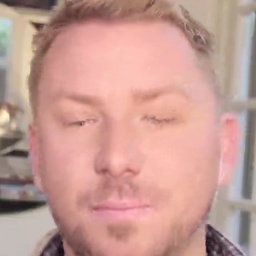} \\
	    
	    	    
	    
	    
	    \includegraphics[width=1.6cm, height=1.6cm]{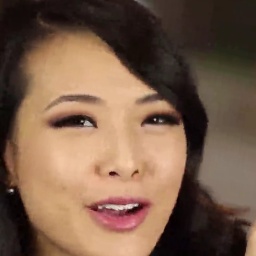} &
	    \includegraphics[width=1.6cm, height=1.6cm]{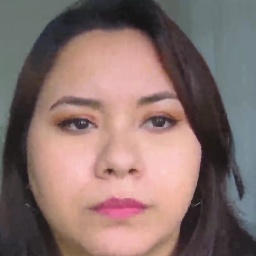} &
	    \includegraphics[width=1.6cm, height=1.6cm]{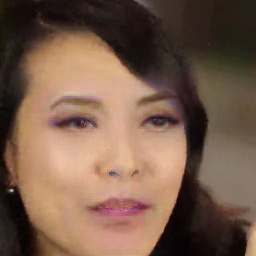} &
	    \includegraphics[width=1.6cm, height=1.6cm]{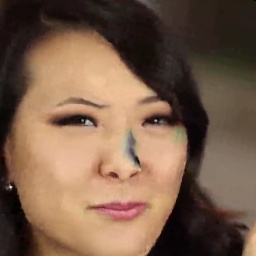} \\
	    
	    
	    \includegraphics[width=1.6cm, height=1.6cm]{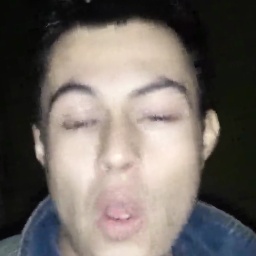} &
	    \includegraphics[width=1.6cm, height=1.6cm]{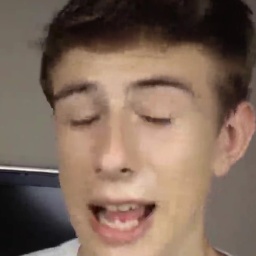} &
	    \includegraphics[width=1.6cm, height=1.6cm]{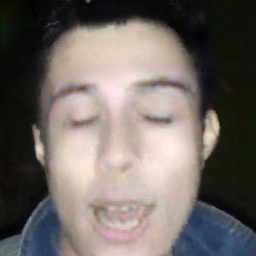} &
	    \includegraphics[width=1.6cm, height=1.6cm]{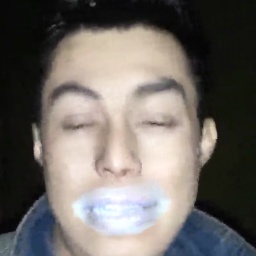} \\
	    
	    


	   
		\bottomrule
	\end{tabular}
	\caption{
    	Qualitative comparison of our proposed model vs Face2Face.
    	(a) - target image;
    	(b) - source image;
    	(c) - Our model;
    	(d) - Face2Face.
	}
	\label{tab-face2face-ablation-img}
\end{table}

\subsection{Results}
We evaluate using \textit{``many-to-many''} (randomly sampling targets and sources) dataset (the model did not ``see'' the images we use here during training).

Ablation studies were executed to compare architectural decisions with possible applications.
Quantitative results of one of them is displayed in Table~\ref{tab-encoders_ablation}.
As we are aiming at better identity preservation and higher realism, we choose separate encoders to siamese ones.

In our decisions, we took into account both quantitative and qualitative \emph{i.e.} visual data (images), albeit not equally.

There is still no ``silver-bullet'' in metrics which could tell definitely that one image will be evaluated better over some other by humans.
During our experiments, we stumbled upon such cases where metrics were good; however, visuals were not. The idea is similar to notorious ``panda-gibbon'' case~\cite{2014arXiv1412.6572G} - generated samples were such to satisfy some metrics while being full of visual artefacts.
These are the main arguments why having both types of data we prioritised visual to numerical results.

\subsection{Forensics}
Nowadays, an average human cannot tell the difference between real and generated face; therefore, it is crucial to have a robust classification model for this task.
To evaluate our results we use close to state-of-the-art model from FaceForensics Benchmark~\cite{roessler2019faceforensicspp} -- Xception c40.
The results are presented in Table~\ref{tab-forensics}. One may see that Xception c40 fails to discriminate on our data.
We acknowledge that for the more fair evaluation fine-tuning of the aforementioned model on our data is required; however, the code for the model training was not released yet.

\begin{table}
    \centering
	\begin{tabular}{llc}
		\toprule
		
		 & & Accuracy of the Xception c40 \\
		\midrule
		
		& & {\textit{``many-to-many''}} \\
		\cmidrule(lr){2-3}
        (a) & & $70 \%$ \\ 
	    (c) & & $25 \%$ \\
	    (d) & & $18 \%$ \\
	    \midrule
		
		& & {face2face dataset} \\
		\cmidrule(lr){2-3}
	    (b) & & $76.7 \%$ \\
	    (d) & & $13.9 \%$ \\
		\bottomrule
	\end{tabular}
	\caption{Classification accuracy of Xception c40.\\
	\mbox{(a) - pix2pix;}
	(b) - face2face method;
	(c) - our with siamese encoders;
	\mbox{(d) - Our with separate encoders.}}
	\label{tab-forensics}
\end{table}

\begin{table}
    \centering
	\begin{tabular}{lcccc}
		\toprule
		
		 & FID $\downarrow$ & NMSE $\downarrow$ & CSIM $\uparrow$ \\
		\midrule
		
		& \multicolumn{3}{c}{\textit{``many-to-many''}} \\ \cmidrule(lr){2-4}
        (a) & $30.77$ & $8.64 \%$ & $0.69$ \\ 
		\addlinespace
	    (b) & \bm{$19.36$} & \bm{$7.13 \%$} & \bm{$0.84$} \\
		\bottomrule
	\end{tabular}
	\caption{Quantitative comparison of face2face vs our proposed model.
	(a)  -  face2face;
	(b)  -  Our model.}
	\label{tab-face2face_ablation}
\end{table}

\subsection{Comparison with others}
In the domain of face-reenactment, it is hard to compare with others due to the lack of published datasets.
We compare with pix2pixHD~\cite{2017arXiv171111585W} and state-of-the-art or close Face2Face~\cite{7780631}, results show advantages of the proposed ActGAN model.
Comparing Pix2pixHD in Table~\ref{tab-pix2pix_ablation} and in Table~\ref{tab-pix2pix-ablation-img}.

We obtained Face2Face dataset from FaceForensics++~\cite{roessler2019faceforensicspp}.
The resolution of videos in FaceForensics++ is quite low, and the faces are in any angles (not frontal) which are additional obstacles for face-reenactment systems.
Results obtained from Face2Face in Table~\ref{tab-face2face-ablation-img} has some visual artefacts (\emph{e.g.} whiteness around the mouth, spots on nose and near eyes), and the expression on generated faces is not corresponding well with the expected \emph{i.e.} source's expression. Whereas the method we propose has better mimics reconstruction, plausible identity preservation and fewer artefacts.

\section{Conclusions}

This paper introduces ActGAN, a powerful and efficient one-shot face reenactment GAN framework, with excellent quantitive and qualitative results.
It advances current works in face synthesis on generation quality and identity preservation.
It also has an efficient, flexible structure and clear training procedure.
We plan to extend ActGAN for real-time face reenactment on videos, and better handling edge-case scenarios. 
We also hope that our research will help researchers to improve the accuracy of the models in the face-forensics domain.

\bibliographystyle{abbrv}
\bibliography{main}

\end{document}